\documentclass[graybox]{svmult}

\usepackage[numbers]{natbib}

\usepackage{mathptmx}       
\usepackage{afterpage}       

\usepackage{helvet}         
\usepackage{courier}        
\usepackage{type1cm}        
\usepackage{makeidx}         
\usepackage{graphicx}        
\usepackage{multicol}        
\usepackage[bottom]{footmisc}
\usepackage{amsmath}
\usepackage{bm}
\usepackage{amssymb}
\usepackage{algorithm}
\usepackage{algpseudocode}
\usepackage{tikz}
\usepackage{listings}
\usepackage{multirow}

\makeindex             

\begin{document}

\title*{A Survey on Artificial Intelligence Trends in Spacecraft Guidance Dynamics and Control}
\titlerunning{A Survey on AI Trends in Spacecraft Dynamics Guidance and Control}
\author{Dario Izzo \and Marcus M\"{a}rtens \and Binfeng Pan}
\institute{Dario Izzo \at European Space Agency, Noordwijk, 2201 AZ, The Netherlands \email{dario.izzo@esa.int}
\and Marcus M\"{a}rtens \at European Space Agency, Noordwijk, 2201 AZ, The Netherlands \email{marcus.maertens@esa.int} \and {Binfeng Pan \at Northwestern Polytechnical University, Xi'an, Shaanxi 710072, China  \email{panbinfeng@nwpu.edu.cn}}
}
%
%
\maketitle

\abstract{
The rapid developments of Artificial Intelligence in the last decade are influencing Aerospace Engineering to a great extent and research in this context is proliferating. We share our observations on the recent developments in the area of Spacecraft Guidance Dynamics and Control, giving selected examples on success stories that have been motivated by mission designs. Our focus is on evolutionary optimisation, tree searches and machine learning, including deep learning and reinforcement learning as the key technologies and drivers for current and future research in the field.
From a high-level perspective, we survey various scenarios for which these approaches have been successfully applied or are under strong scientific investigation. Whenever possible, we highlight the relations and synergies that can be obtained by combining different techniques and projects towards future domains for which newly emerging artificial intelligence techniques are expected to become game changers.}

\section{Introduction}
\label{sec:1}
The impact of Artificial Intelligence (AI) technologies on our modern world is now more apparent then ever.
Once a niche research field in computer science and mathematics, the vision of machines displaying human levels of intelligence has spread and matured into solid and well established industrial products.
Powerful techniques from Machine Learning (ML) have gained increasing attention from scientists over the past few decades and are now transforming whole economies as a result.

The space sector is catching up to these developments, as more and more works are published that incorporate concepts related to AI, like natural language processing, knowledge representation, automated reasoning, computer vision, robotics, etc. Applications of interest range from preliminary spacecraft design to mission ope\-ra\-tions, from guidance and control algorithms over navigation to the prediction of the dynamics of perturbed motion and towards classification of astronomical objects and refinement of remote sensing data to only name a few.

The goal of this survey is to present a slice of this work to highlight the progress that has been made in the adoption of AI techniques. More precisely, our focus will be on the recent AI trends that have emerged in spacecraft guidance dynamics and control. Even when limited to this area, a comprehensive overview of all work would quickly grow out of proportions, which is why we can only present selected pointers to the reader. We decided to give importance to work published in the last few years, avoiding the historical perspective of older and well-established fundamental works. Additionally, we decided to avoid publications which are strongly speculative in nature: while visionary ideas are interesting to follow, an important requirement for this survey was a well-motivated applicability for a space-related challenge, ideally inspired by an already established or newly proposed mission concept.
This narrow scope allows our survey to be concise while remaining relevant for the interested practitioner.

Many times, results obtained by one AI technology for a specific task appear stunning, but perform rather poorly when transferred to a different task, which often happens when its strength and weaknesses are not thoroughly understood. However, due to the pioneering works of many researchers combined with the results of large competitions, like the Global Trajectory Optimisation Competition (GTOC), acting as benchmarks, a better understanding about the most promising approaches has been obtained. In particular, this survey will guide the reader through the intersection of Guidance and Control with evolutionary optimisation, tree searches and machine learning (including Deep Learning and Reinforcement Learning) for which we witnessed a stream of results of remarkable quality. Consequently, we devote one section to each of the aforementioned techniques in that order.
Additionally, we contribute by highlighting the synergies and relations between these techniques, as it has been frequently demonstrated that they can benefit vastly from each other if combined, which is often the most viable solution strategy to achieve optimal results for the complex challenges in space.

\subsection{Related Surveys}

We believe that our report on the state of the art in AI for Guidance and Control is timely and, to the best of our knowledge, has not been reported in this form yet by others. However, there exists work with a different scope in close proximity to our area which might be useful to deepen or broaden ones knowledge for selected topics. Girimonte and Izzo~\cite{girimonte} give a general overview focused on distributed AI for swarm autonomy and distributed computing for enhanced situation self-awareness and for decision support in spacecraft system design. Some useful examples of applied machine learning for geoscience and remote sensing are given by Lary~\cite{lary2010artificial}. A more recent review by Zhu et al.~\cite{Zhu2017Deep} highlights recent advances in remote-sensing data analysis with a main focus on Deep Learning in particular. For a stronger historical perspective on the development of evolutionary optimisation and machine learning over time in trajectory optimisation, we suggest to consult Izzo et al.~\cite{izzo2018machine}. A survey on the methods applied by the international community during the GTOC (and their Chinese equivalent CTOC) was just recently published by Li et al.~\cite{gtocreview} and contains major contributions from AI algorithms and methodologies. Lastly, the book from Russel and Norvig~\cite{russell2010artificial} provides a good starting point to obtain a comprehensive overview about modern AI detached from a particular application domain.  

\subsection{Glossary}

While this survey explains its acronyms along the way, we provide the following list of reoccurring abbreviations as a reading aid:\\[6pt]
\begin{center}
\begin{tabular}{p{1pt}p{2.3cm}l}
\hline 
 & $\circ$ AI & Artificial Intelligence  \\ 
 & $\circ$ ACO & Ant Colony Optimisation \\ 
 & $\circ$ ANN & Artificial Neural Network\\
 & $\circ$ CMA-ES & Covariance Matrix Evolutionary Strategy  \\
 & $\circ$ CNN & Convolutional Neural Network  \\ 
 & $\circ$ CTOC & Chinese Trajectory Optimisation Competition \\ 
 & $\circ$ DE & Differential Evolution \\
 & $\circ$ DL & Deep Learning \\  
 & $\circ$ DNN & Deep Neural Network \\  
 & $\circ$ DQN & Deep Q-Learning Network \\ 
 & $\circ$ DRL & Deep Reinforcement Learning\\
 & $\circ$ GA & Genetic Algorithm \\ 
 & $\circ$ G\&CNETS & Guidance and Control Networks \\ 
 & $\circ$ GTOC & Global Trajectory Optimisation Competition \\ 
 & $\circ$ GTOP & Global Trajectory Optimisation Problem \\
 & $\circ$ LSTM & Long Short-Term Memory \\ 
 & $\circ$ MCTS & Monte Carlo Tree Search \\   
 & $\circ$ ML & Machine Learning \\
 & $\circ$ NSGA-II & Non-Dominated Sorting Genetic Algorithm (Mark 2) \\  
 & $\circ$ PSO & Particle Swarm Optimisation \\ 
  & $\circ$ RL & Reinforcement Learning \\
 & $\circ$ SVM & Support Vector Machine \\ 
\hline 
\end{tabular} 
\end{center}

\section{Evolutionary Optimisation}
\label{sec:2}

Evolutionary algorithms are a class of global optimisation techniques that make use of heuristic rules, often inspired but not limited to natural paradigms such as Darwinian evolution. For example, a standard Genetic Algorithm (GA) encodes a po\-pu\-la\-tion of solutions, which undergoes mutation, crossover and selection to search for (close to) optimal solutions in often heavily discontinuous and rugged landscapes. As such, GAs have proven to be very useful to solve interplanetary trajectory optimisation problems where the planetary constellations define a complex solution landscape, exhibiting multiple close-by minima already in simple cases such as that of planet to planet transfers. 

The ecosystem of genetic and evolutionary algorithm (sometimes referred to as meta-heuristics) together with their variants and mixtures is so vast that it seems almost futile to summarize. The biological inspirations for these algorithms range from insect swarming, foraging and ant behaviour, to even more obscure examples including algae, the American buffalo, humpback whales and penguins. A rather exhaustive list of this evolutionary computing bestiary is maintained by Campelo et al.~\cite{bestiary}. Nevertheless, certain types of algorithms have been deployed consistently and with high degrees of success to solve challenges in aerospace.
Most notably, trajectory optimisation problems, such as the numerous ones assembled in the Global Trajectory Optimisation Problem (GTOP) database~\cite{vinko2008global} (a sort of open trajectory optimisation gym), are frequently solved and studied with such methodologies~\cite{stracquadanio2011design, addis2011global, schlueter2014midaco, islam2012adaptive, cassioli2012machine, simoes2014self}, building a domain-specific benchmark for performance of those evolutionary techniques. Some of the interplanetary trajectory problems (e.g. Messenger and Cassini2) in the GTOP database were also used during the CEC2011 competition, attracting the attention of the larger scientific community of scientists involved in evolutionary computations to work on space-related problems (see~\cite{elsayed2011ga} for the competition winner).

In the following, we discuss some of the algorithms and approaches ordered by their applicability to different forms of optimisation problems, beginning with single-objective unconstrained and leading to multi-objective and combinatorial constrained problems.

\subsection{Single-objective, Unconstrained, Continuous Problems}
Differential Evolution (DE) is a comparatively simple variant of a GA that is effective for nonlinear and non-differentiable continuous space functions as encountered frequently, for example, in chemical propulsion spacecraft transfers where sequences of multiple impulsive velocity increments need to be decided. After Myatt et al.~\cite{ari1, izzo2007search} introduced its use in this context, Olds and Kluever~\cite{olds2007interplanetary} made an analysis of the DE's performance with respect to the Cassini and Galileo missions (among others) finding the expected sensitivity of the algorithm performance on its parameters. Recent research in DE is thus concerned with self-adaptation, i.e. the incorporation of hyper-parameters like the mutation rate into the chromosome such that they are evolved together with the optimal solution. Izzo et al.~\cite{izzo2013search} deploy a self-adaptive DE to design a grand tour between the Galilean moons of the Jovian system. Yao et al.~\cite{yao2017improved} propose a double self-adaptive DE with random mutations and evaluate its performance on a Lambert transfer problem. Theoretical advances on the DE algorithm itself have also been obtained by Vasile et al.~\cite{vasile2011inflationary} who were able to use theoretical insights on DE's evolutionary mechanisms to design an algorithm able to outperform, on some trajectory optimisation problems, a canonical (i.e. not self-adapted) DE variant.

Similar in popularity to DE is Particle Swarm Optimisation (PSO), a bio-inspired search-heuristic with clear links to the foraging behaviour of flocks of birds, school of fishes or similar types of intelligent swarms. Pontani and Conway~\cite{pontani2010particle} give a comprehensive overview about the adaptation of this technique towards spacecraft trajectory optimisation. Benefits of PSO are its comparatively easy implementation and a generally high convergence speed to global optima with good accuracy. Vasile et al.~\cite{vasile2010analysis} benchmark PSO, DE and other evolutionary algorithms against each other for different setups, highlighting the problem dependency of meta-heuristic performances (a general issue in meta-heuristics known as the ``no free lunch theorem''~\cite{wolpert1997no}). It is thus advisable in practice, to not rely too heavily on a single meta-heuristic, but to explore their performances in parallel or to even combine several meta-heuristics into one. Englander and Conway~\cite{Englander2012Automated} achieve their best performance by a combination of DE and PSO for an automated mission design for sequences of interplanetary transfers including multiple gravity assist maneuvers. Sentinella and Casalino~\cite{sentinella2009hybrid} deploy PSO, DE and other GAs together to obtain the global optimal solution consistently for Earth-Mars transfer scenarios.

Another, more sophisticated evolutionary meta-heuristic that was recently investigated in the context of trajectory optimisation by Izzo et al.~\cite{izzo2014constraint} is Covariance Matrix Evolutionary Strategy (CMA-ES). CMA-ES is a GA that deploys an adaptive mutation scheme, exploiting the pairwise dependencies of the decision variables given by their covariance matrix. The basic idea is to increase the likelihood of selections that have been proven beneficial before. While the implementation details of CMA-ES are much more involved than DE or PSO, it has been shown to be able to outperform these approaches on a large class of interplanetary transfer problems. 

Last but not least, Radice et al.~\cite{radice2006ant} analyze Ant Colony Optimisation (ACO) for an Earth-Mars transfer inspired by the Mars Express mission. The ACO paradigm simulates the foraging behaviour of natural ant colonies, where ants deposit biomarkers (pheromones) along their paths to communicate and further reinforce exploration in comparatively large search environments. While ACO is traditionally deployed in discrete domains, the authors show how the algorithm can be modified for continuous optimisation problems (even though their proposal seems to be only a preliminary attempt with large margins for improvement). MIDACO developed by Schlueter et al.~\cite{schlueter2013midaco, schlueter2014midaco} is an optimisation framework that deploys ACO to problems that can be single or multi-objective, allowing for constrained and mixed integer decision variables as well. Some of the best solutions to trajectory problems in the GTOP database, although single-objective (i.e. Cassini2, GTOC1, MessengerFull), were discovered by MIDACO.

\subsection{Multi-objective Problems}

The researches mentioned in the previous section are mostly concerned with the optimisation of box constrained continuous variables towards a single objective, most commonly fuel consumption or total travel time in the case of interplanetary trajectories. However, the true nature of most design problems is multi-objective, potentially also including integer decisions variables or non-linear constraints. In this multi-objective setting, the concept of the best design (i.e. the global optimum) is substituted with that of a Pareto front, a collection of non-dominated solutions expressing the trade-offs between different conflicting objectives. Consequently, a set of best possible solutions (Pareto-optimal front) is required to guide engineering decisions (i.e. trajectories, who could have been improved in one objective without sacrificing another). Since in this context we are interested in a set of solutions, population-based algorithms such as evolutionary algorithms are a natural choice offering substantial advantages. Indeed, multi-objective problems are one of the prime examples for the success of evolutionary algorithms in general, for which they are the de facto standard approach (see Coello~\cite{coello2006evolutionary}).

One of the classic evolutionary approaches to multi-objective optimisation is the Non-Dominated Sorting Genetic Algorithm (NSGA-II), which has been studied for the optimisation of planetary fly-by sequences including integer variables also by its inventor (see Deb et al.~\cite{deb2007interplanetary}). Schutze~\cite{schutze2009designing} considers a bi-objective approach for the design of multiple low-thrust gravity assist trajectories (minimisation of flight time and fuel consumption) and deploys NSGA-II on that landscape. 
Later studies by M\"{a}rtens and Izzo~\cite{Martens2013Asynchronous} show how the performance of NSGA-II can be improved by implementing a multi-objective migration scheme in an island model. The authors evaluate their approach for a transfer from Earth to Jupiter, including multiple approximate Pareto-optimal fronts for different fly-by sequences. Similar trajectories are also under investigation by Zotes and Penas~\cite{Zotes2012Particle}, who deploy a multi-ojective extension of PSO called MOPSO. Lee Seungwon et al.\cite{lee2005multi} benchmark several evolutionary multi-objective techniques, including NSGA-II, to find the non-dominated front depending on the free control parameters of a Q-law feedback controller.

Izzo et al.~\cite{izzo2014constraint} analyse a decomposition-based approach to multi-objective optimisation named MOEA/D (Multi Objective Evolutionary Algorithm by Decomposition), which is shown to improve NSGA-II's approximation of the Pareto-optimal front significantly for transfer problems in the Jovian system. Furthermore, different constraint handling techniques like co-evolution and artificial immune systems are applied and benchmarked. A rather comprehensive survey on multi-objective methods for spacecraft design can be found in Montano et al.~\cite{montano2014multi} where several evolutionary approaches are described and discussed.

Besides trajectory optimisation, also other guidance and control problems have been analysed in a multi-objective setup. Vasile and Ricciardi~\cite{Vasile2016direct} introduce a memetic multi-objective algorithm to solve the optimal control problem. The approach is evaluated on two tasks: a rocket launch trajectory for which minimum time and maximum horizontal velocity are optimized and an orbit rising task that optimizes final energy and maneuver time. Chai et al. \cite{chai2018unified}, discuss an aero-assisted trajectory optimisation problem with mission priority constraints.
They propose a gradient-based hybrid GA surpassing the need to perform non-dominated sorting (required by algorithms like NSGA-II) and thus is more efficient in contexts where the complexity of the non-dominated sorting operations becomes a practical issue.

\subsection{Combinatorial Problems}

With an increasing complexity, interplanetary trajectories can no longer (effectively) be described by continuous and unconstrained decision variables alone. For example, if multiple gravity assisted fly-bys become imperative, the sequence of planetary encounters is part of the optimisation problem and results in a combinatorial dimension that often turns out to be crucial for achieving the mission goals.
A similar combinatorial part also appears in many other mission profiles where multiple bodies or rendezvous points have to be considered. The ninth edition of the global trajectory optimisation competition (GTOC9) asked for a debris removal mission in which a large number of space debris need to be visited (and deorbited) in quick succession while minimizing the amount of launches necessary (see Izzo and M\"{a}rtens~\cite{Izzo2018Kessler} for details on the development of the challenge). In this situation, the aforementioned ACO has proven to be particularly effective. In fact, the top scoring teams~\cite{gtoc9JPL, gtoc9NUDT, gtoc9XSCC} in the GTOC9 made use of some ACO variant. Furthermore, a variant of ACO is also deployed by Ceriotti and Vasile~\cite{ceriotti2010mga} to optimize missions containing multiple fly-by maneuvers like Cassini and Laplace. 

The full automation of the interplanetary trajectory pipeline is also explicitly proposed by Englander et al.~\cite{englander2016automated}, who makes use of a number of well assembled techniques to propose an automated solution strategy. In this work, the combinatorial part of the problem is solved by a GA while an approach based on monotonic basin hopping takes care of the inner low-thrust problem as suggested in the work of Yam et al.~\cite{yammbh}.

Other challenges worth mentioning include mission concepts inside the asteroid belt where the number of astronomical bodies and the sequence of possible transfers lead to a combinatorial explosion as multiple categorical decision variables have to be considered. Izzo et al.~\cite{izzo2014gtoc5} design a GA for a multiple asteroid rendezvous mission which allows to deactivate (or hide) certain genes in the solution representation. The systematic use of hidden genes in spacecraft trajectory optimisation is further studied by Abdelkhalik and Darani~\cite{hiddenOSSAMA} who design a Jupiter transfer including the optimisation of intermediate fly-by using this technique.

Some particularly complex interplanetary trajectory optimisation problems, like it is the case for active multiple debris removal missions, can also be interpreted as Travelling Salesman Problems and be dealt with accordingly. Izzo et al.~\cite{tsp_izzo} show that once the problem is framed as a Travelling Salesman Problem, specific heuristics like inver-over become effective operators for evolutionary algorithms.

\section{Tree Searches}
\label{sec:ts}

Although direct application of evolutionary algorithms to combinatorial problems is possible, it may lead to sub-optimal results if the search space becomes too large to be sampled effectively. An alternative approach to deal with these problems is that of deploying tree searches, which becomes an option whenever the problem domain allows to construct a solution incrementally by the sub-solutions to smaller, separable sub-problems. This is often the case for the combinatorial challenges mentioned before, i.e. complex rendezvous, fly-by problems, etc. which can be seen as bi-level optimisation problems, in which the combinatorial selections at the outer level influence the resulting fitness landscape of the inner continuous level. Vice versa, the performance of the optimized trajectories at the inner-level feed back into guiding the selection on the outer level.

Tree search methods emerged within these contexts as one of the most successful approaches for this type of challenge. In a tree search, decision points (i.e. which orbital body should the spacecraft visit next?) are modeled as nodes which can be expanded for further evaluation. As an exhaustive enumeration of all possible node expansions becomes quickly intractable, each tree search deploys a strategy that only explores the most promising branches as needed and produces sub-problems that can be handled by evolutionary algorithms more easily.

Wilt et al.~\cite{Wilt2010comparison} study such strategies on multiple benchmark problems (although none inspired by interplanetary mission designs). They conclude that the strategy of the Beam Search algorithm is the best choice, especially when confronted with massive search spaces. This might be the reason, why the Beam Search algorithm emerged in many works for preliminary studies of complex missions.

Arguably one of the best features of Beam Search is its balance between the exploration and exploitation of the search tree by limiting its expansion to a subset of nodes determined by a ranking criteria. For example, Lazy-Race Tree Search (see Izzo et al.~\cite{izzo2013search}) is a Beam Search that ranks the (partial) trajectories by their total time of flight while each node is expanded to minimize mass consumption. This allows for an effective exploration on different levels of the tree in comparison to a greedy search strategy which would, typically, proceed level-wise through the tree selecting only the best few solutions and discarding path that would have become promising to explore at deeper levels.

While ranking criteria are typically deterministic, expansion of the search tree can also be guided by a stochastic process.
Hennes and Izzo~\cite{hennes2015interplanetary} study Monte Carlo Tree Search (MCTS), a tree search variant frequently applied to develop agents for difficult games with large search spaces like the game of Go (Compare Browne et al.~\cite{Browne2012Survey}). By applying MCTS to the large design space of the Cassini missions, the authors were able to rediscover the correct sequence of planetary encounters within a time line that was very close to the actual trajectory flown.

Another stochastic tree search that hybridizes Beam Search with ACO is Beam P-ACO as introduced by Sim{\~o}es et al.~\cite{simoes2017multi}. Following this approach, the search paths of the tree are modified by pheromone markers to reinforce exploitation of promising paths. Furthermore, Beam P-ACO is implemented as an anytime algorithm and can thus provide a solution at every point during its execution, allowing for an easy fine-tuning between computational resources and solution quality. The winning team in the GTOC9 competition, from NASA's Jet Propulsion Laboratory, made use of a modified Beam P-ACO algorithm as part of their solution strategy~\cite{gtoc9JPL}.

\section{Machine Learning}
\label{sec:ml}

While evolutionary techniques are widely adopted to aid the design of spacecraft trajectories, the use of machine learning (ML) concepts like Artificial Neural Networks~(ANNs), Support Vector Machines (SVMs), Decision Trees, Random Forests, etc. is still at its beginning. Reasons for the slow adoption of ML include the lack of publicly available and suitable large scale data sets for aerospace challenges and the sometimes less obvious applicability of these methods to the typical problems encountered. Nevertheless, the interest in ML is high and more research is produced every year that features, in one form or another, a machine learning model. In the following, we first highlight some important connections that exists between ML and evolutionary algorithms, before we focus on Deep Learning~(DL) as one of the major technologies. To complement DL, we briefly mention some alternative approaches for supervised learning before concluding with important work that applies and explores the reinforcement learning paradigm.

\subsection{Machine Learning and Evolutionary Optimisation}

ML algorithms have many relations to the evolutionary techniques discussed before. The most common relation is the generation of training data: during the optimisation of an interplanetary transfer, as in any optimisation task, a large number of solutions are computed and assessed to inform the search for better candidates. These design points are used to guide the search into more promising directions, but are typically discarded due to their intermediary nature.
However, these (partial) trajectories and solutions become extremely valuable when applied within a supervised learning setup. In particular, future evolutionary runs may improve significantly if initial guesses are provided by a ML model as demonstrated by Cassioli et al.~\cite{cassioli2012machine} on some of the trajectory problems in the GTOP database by using SVMs.

Similarly, Basu et al.~\cite{Base2017Timeoptimal} study the design of time-optimal slew maneuver for a rigid space telescope in order to reorient some light sensitive parts to observe quick events like gamma ray bursts. While time-optimal solutions can be obtained with PSO, the convergence time of the swarm is considered too high to be practical for this task. As a solution, a neural network is deployed to predict advantageous initial conditions for the PSO, significantly reducing the convergence time and moving this approach closer to a real-time optimal control system.

Machine learning may also construct an inexpensive proxy to the objective function (Ampatzis et al.~\cite{ampatzis2009machine}) if trained with the design points sampled during evolutionary runs. Building such a surrogate model is particularly relevant if the computation of some interplanetary mission fitness requires a high amount of computational resources and effort such as in the case of optimal low-thrust transfers. In this case, a surrogate model approximating the final optimal transfer mass enables to quickly search for ideal launch and arrival epochs, as well as favorable planetary body sequences, e.g. in the case of multiple asteroid or debris rendezvous missions as shown in the works of Hennes et al.~\cite{hennes2016fast} and Mereta et al.~\cite{mereta2017machine}.

Unsupervised learning techniques such as clustering or nearest neighbours have also been deployed to select the target of transfers in multiple asteroid rendezvous missions, upon proper definition of a metric coping with the orbital non-linearities (Izzo et al.~\cite{izzo2016designing}), or to define new box bounds and hence focus successive evolutionary runs in promising areas of the search space by cluster pruning (Izzo et al.~\cite{izzo2007search, izzo2010global}).

Lastly, GAs have been used to directly modify the weights of a single layer neural network (neuroevolution) in the work of Dachwald~\cite{dachwald2004low,bernd:2018} for low thrust transfers. However, most recently, deep neural networks have shown superior performance in this setup and provide a promise of future on-board real-time computation of guidance profiles. In the following subsection, we highlight recent work that follows this paradigm and can be classified as DL.

\subsection{Deep Learning}

Artificial Neural Networks provide the most popular and successful applications in the field of machine learning in our decade. In particular, Deep Neural Networks (DNN), i.e. networks with a large number of hidden layers, are frequently deployed to learn a model from a database of examples (see Schmidhuber~\cite{schmidhuber2015deep}) using some form of gradient descent.

S\'{a}nchez-S\'{a}nchez et al.~\cite{sanchez2016learning, sanchez2016real} provide a systematic study on how DNNs can be trained on optimal state feedback of continuous time, deterministic, non-linear systems like inverted pendulum stabilization, pinpoint landing of a multicopter and landing of a spacecraft (imitation learning or supervised learning).
These systems include cost functions for smooth continuous, but also discontinuous (bang-bang) optimal control, as is for example the mass optimal case considered there.
It is shown that a deep network is able to learn all tasks with remarkable accuracy and, for the multicopter and spacecraft scenario, that the network generalizes well outside the bounds of the training set whenever deep network topologies are deployed. The authors also suggest that deep architectures do more than merely interpolate between data but may actually learn underlying dynamic principles of the models under investigation, like the Hamilton-Jacobi-Bellman equations. Izzo et al.~\cite{izzo2018machine} extend these results to interplanetary trajectories, showing how it is possible to train a DNN to guide a spacecraft optimally from an Earth orbit to a Mars orbit. The same authors~\cite{izzo2018neurostability} also propose the name Guidance and Control Networks (G\&CNETs) to indicate a generic deep architecture trained to perform optimal manoeuvres using the imitation learning (or supervised learning) paradigm, and provide a new method based on differential algebra and automated differentiation to study their stability margins and controlling performances. As such, G\&CNETs are one of the most promising Deep Learning based technologies that can potentially simplify the on board control and guidance software replacing it with one, relatively simple, trained neural model. While G\&CNETs are simple feed forward networks, more complex, recurrent topologies are explored as well, i.e. Recurrent Neural Networks based on Long Short-Term Memory (LSTM) by Furfaro et al.~\cite{furfaro2018recurrent} in a similar context.

Since accurate position information might not be readily available due to technical limitations and costs of sensory hardware, there is a recent trend to develop control networks solely based on simple visual cues and optical flow that can be obtained by comparatively inexpensive cameras (see Franceschini~\cite{franceschini2014small}). The most popular architectures for image processing within the Deep Learning world are arguably Convolutional Neural Networks (CNNs) due to their superior performance in image classification benchmarks~\cite{krizhevsky2012imagenet}.
In a different work of Furfaro et al.~\cite{Furfaro2018Deep}, a stack of DNNs is assembled, starting with a CNN for the visual processing of simulated moon images and followed by a Recurrent Neural Network to estimate landing controls in a 2D simulated environment. The Recurrent Neural Network is based on LSTM cells, which are a known for their good performance on sequential or time-dependent data. The total network consists of almost 30M trainable parameters and performs classification on the thrust vector magnitude (bang-bang) and regression on the thrust vector angle, both with high degrees of accuracy.



\subsection{Alternative Supervised Learning Approaches}

Acquisition of reliable data from space remains one of the biggest roadblocks for DL in general. However, more parsimonious alternatives to DNNs are under study as well: Shang et al.~\cite{Shang2018Parameter} design fuel and time optimal trajectories for transfers between all asteroids in the inner belt (around 150K) by a Gaussian Process Regression model.
The challenge is to accurately predict the orbital parameters of the corresponding pairwise transfers which would take a considerable amount of time to compute by conventional methods.
By grouping asteroids of similar orbits together, sophisticated feature engineering allows to train the Gaussian Process Regression model on merely 300 numerically derived training samples to reach high levels of accuracy.

Shah and Beeson~\cite{Shah2017Vishwa} apply neural networks and random forests (among GAs) to approximate manifold structures emerging in three-body problems. The authors compare their trajectories with cubic convolution, the current state-of-the-art method for approximation of such manifolds and find that Random Forests perform reasonably well across multiple orbital energies.

Support Vector Machines (SVMs) provide an alternative supervised learning approach for both, classification and regression problems. SVMs can efficiently handle linear and nonlinear problems, relying on the kernel functions which are used for training. Another key property of SVMs is their universal approximation capability with various kernels, including Gaussian, several dot product, or polynomial kernels~\cite{Hammer2003A}.

Li et al.~\cite{Li2015Trajectory} use SVMs to classify the trajectories in the circular restricted three-body problem. Based on the SVM approach, the transit orbits, which can pass through the bottleneck region of the zero velocity curve and escape from the vicinity of the primary or the secondary, can be rapidly separated from the other types of orbits. 
Peng and Bai~\cite{peng2018exploring} explore the capabilities of SVMs for improving orbit prediction accuracy. The SVM model is designed and trained to learn the underlying pattern of the orbit prediction errors from historical data. The simulation results demonstrate that the trained SVM model is able to capture the underlying relationship between the learning variables and the desired orbit prediction error. However limitations to its generalization apply if predicted orbits are too far in the future~\cite{peng2017limits}. The authors later also deploy ANNs on the same problem~\cite{Peng2018Artificial} complementing the SVM approach.

\subsection{Reinforcement Learning}

Reinforcement Learning~(RL) is a sub-discipline of machine learning in which agents learn how to solve a task (like navigation or planning) within a potentially changing environment. The system receives feedback by means of a reward function, which informs the agents how good or how bad they are currently performing. The overall goal is to maximize the cumulative reward of this function by selecting the best (sequence) of actions given certain observations of the environment. An advantage of RL is its ability to adapt to unknown situations and circumstances that are difficult to foresee or cumbersome to account for manually. Consequently, challenges in autonomous navigation and control are natural fits for RL as soon as uncertainty arises and robustness is of importance.

Typical challenges involve controllers that enable spacecrafts to hover and orbit irregular shaped bodies (like asteroids). First explorations by Gaudet and Furfaro~\cite{Gaudet2012Robust} show that RL is able to learn the non-uniform gravitational and rotational fields of simulated asteroids to develop a thrust-profile for accurate and robust hovering. This task would otherwise be extremely time-consuming to solve with traditional methods given the plethora of differently shaped asteroids found just in our solar system.

Willis et al.~\cite{Willis2016reinforcement} improve upon the previous work by increasing the accuracy of the approach by an order of magnitude and transferring the task to a more general gravitational model. A further improvement is the transfer to optical flow as single source of sensory information for the controller. Interestingly, the weights of the network deployed for this scenario are trained with neuroevolution by a variant of PSO.

There is an additional stream of work emerging outside the terminology of machine learning, but strongly related to RL. Most notably, Pellegrini et al.~\cite{Pellegrini2017Multiple} and Ozaki et al.~\cite{Ozaki2015differential} explore dynamic programming for low thrust transfers. Also here, the main motivation is to provide a robust optimal control in uncertain environments.

Looking at AI in general, arguably one of the most effective approaches in this decade was the arrival of Deep Reinforcement Learning~(DRL), spearheaded by the success of AlphaGo~\cite{Silver2016mastering, Silver2017mastering}, the first neural network capable of beating human world champions at the game.
While the first iteration (AlphaGo) learned from a database recording several human matches of Go, later variants (like AlphaGo Zero) made heavy use of competitive self-play, i.e. training with previous iterations of itself, starting from random actions.
This creates a large amount of data, on which the system gradually improves itself. One of the key ingredients of the overwhelming success of this approach is experience replay, a history of previously learned action and reward pairs that is being replayed during training.

So far, adaptation of DRL in the domain of Guidance and Control is slow, but first studies are emerging: Chu et al.~\cite{Chu2018Q} deploy a Deep Q-learning Network (DQN) to design a rendezvous-mission, where one satellite needs to meet a second satellite within a constellation while avoiding collisions. The deployed Q-Learning algorithm allows the satellite controller to make sequential decisions on the path to take, safely navigating the environment. As this problem features extremely high-dimensional state and action spaces, the standard Q-Learning algorithm had to be augmented by a DNN, resulting in a DQN. By deploying the DQN to approximate the high-dimensional reward-function, the satellite controller is able to produce two-key behaviours necessary to complete the task: obstacle avoidance and target seeking. Most notably, experience replay is deployed in this setting as a mechanism to prevent divergence and stabilize the training of the DQN.

Very recently, Gaudet et al.~\cite{gaudet2018deep} published a work on Deep Reinforcement Learning on a planetary descent and landing problem, which deploys Policy Gradient Optimization as the learning algorithm. The authors design and train a 6-degree-of-freedom integrated closed-loop controller capable of fuel-optimal pinpoint landing, satisfying varying flight and system constraints.
The work highlights the importance of shaping the reward function towards the task and provides exhaustive technical details on how the DNN is trained for this setting.


\section{Final Remarks}
\label{sec:4}

The use of results and ideas originated in the AI research community has been already fruitful in domains such as orbital prediction, planetary landing, spacecraft guidance, interplanetary trajectory optimisation and low-thrust propulsion, allowing the development of novel methods and architectures that are competitive, and often superior, to the state of the art methodologies currently used and taught to aerospace engineers. We are confident that a widespread use of these techniques, currently only known and used by academics and a few practitioners, will increase the level of automation and the performances of space systems already in the next decade. This trend has already begun and is continuing, taking now advantage of the increased attention on AI that happened worldwide thanks to the success of paradigms such as Deep Learning in the IT industry.

We expect that more success stories in new domains such as formation flying, rendezvous and docking, in-orbit self-assembly, or autonomous detection on planet/asteroid surface will appear in the next decades with Deep Learning and Deep Reinforcement Learning powering many developments in all areas as per their novelty and increasing attractiveness.
Work on the validation of AI based system is also expected to appear to provide methods for increasing the trust in trained models (like DNNs), which are often regarded with skepticism when seen as a black box.


\bibliographystyle{astrobib}
\bibliography{main}
\end{document}